\documentclass[sigconf,nonacm]{acmart}

\AtBeginDocument{%
  }

\setcopyright{acmcopyright}
\copyrightyear{2023}
\acmYear{2023}
\acmDOI{XXXXXXX.XXXXXXX}

\acmConference[ICMI '23]{25th ACM International Conference on Multimodal Interaction}{October 09--13, 2023}{Paris, France}
\acmPrice{15.00}
\acmISBN{978-1-4503-XXXX-X/18/06}

\newcommand{\E}{\mathbb{E}}
\newcommand{\pr}{\mathbb{P}}
\usepackage[ruled,vlined]{algorithm2e}
\newcommand{\R}{\mathbb{R}}
\DeclareMathOperator*{\argmax}{arg\,max}

\begin{document}

\title{Early Classifying Multimodal Sequences}

\author{Alexander Cao}
\affiliation{%
  \institution{Northwestern University}
  \city{Evanston}
    \state{Illinois}
  \country{USA}
}
\email{a-cao@u.northwestern.edu}

\author{Jean Utke}
\affiliation{%
  \institution{Allstate Insurance Company}
  \city{Northbrook}
    \state{Illinois}
  \country{USA}
}
\email{jutke@allstate.com}

\author{Diego Klabjan}
\affiliation{%
  \institution{Northwestern University}
  \city{Evanston}
    \state{Illinois}
  \country{USA}
}
\email{d-klabjan@northwestern.edu}


\begin{abstract}
Often pieces of information are received sequentially over time. When did one collect enough such pieces to classify? Trading wait time for decision certainty leads to early classification problems that have recently gained attention as a means of adapting classification to more dynamic environments. However, so far results have been limited to unimodal sequences. In this pilot study, we expand into early classifying multimodal sequences by combining existing methods. We show our new method yields experimental AUC advantages of up to 8.7\%.
\end{abstract}

%



\maketitle

\section{Introduction}
Early classifying multimodal sequences is ubiquitous in our lives. Scanning through movies on any streaming platform, the trailer begins to play. Based on seconds of video and audio, we make a quick decision of whether or not to watch it. A more serious example is a physician diagnosing a patient. From imaging, lab tests, and other results arriving at different times, the doctor is attempting to diagnose as accurately as possible (to start the correct treatment) as quickly as possible (to begin that treatment sooner). With such applications, it is important to adapt early classification methods to multimodal sequences as they exhibit their own challenges.

Specifically, early classification manifests in the following problem setup: At each time step we receive new information in a sequence. From here, we must decide if we have enough information to stop and predict the class with sufficient accuracy, or if we should continue waiting for additional information in later time steps to improve prediction accuracy. Balancing this dual objective of classifying as soon as possible, as accurately as possible is the core problem.

To solve this, we combine an OmniNet-like \cite{pramanik2019omninet} transformer neural network with Classifier-Induced Stopping (CIS) \cite{cao2023policy}. Transformers \cite{vaswani2017attention} represent the state-of-the-art neural network for sequence-based tasks and OmniNet further advances them by explicitly modeling spatial-temporal interaction, making its architecture well-suited for our early classification of multimodal sequences. The recent CIS provides an efficient method to learn both a policy for deciding between stopping and waiting and a classifier. It works by finding the optimal stopping time based from its own classifications each time step. 

Our contributions are two-fold. First, this is a pilot study in early classification of multimodal sequences. To our knowledge, this is first time early classifiers have been applied to sequences composed of different modalities such as images, text, and structured categorical data. Second, we demonstrate that spatial-temporal transformers in combination with CIS is a potent model for early classifying multimodal sequences. Experiments show our method holistically outperforms a similar benchmark early classifier with the same neural network body. 

This paper is structured as follows. In \S \ref{sec:related}, we review related work in multimodal neural networks and early classifiers. In addition, we introduce relevant notation and detail the benchmark method. In \S \ref{sec:omninet} and \S \ref{sec:cis}, we thoroughly lay out OmniNet's spatial-temporal transformer and CIS, respectively. All experimental details and results are explained in \S \ref{sec:exp}. Finally, we conclude in \S \ref{sec:conclude}. 

\section{Related work} \label{sec:related}

\subsection{Neural Networks for Multimodal Sequences}
There is a large body of work adapting LSTMs \cite{hochreiter1997long} to digest multimodal sequences. These methods can be distinguished by the stage of fusing of the different modalities. \cite{tang2017multimodal} concatenates features from different modalities and feeds this larger input into an LSTM. \cite{agarwal2019multimodal, feng2017audio} reserve separate LSTMs for each modality and then fuse the outputs. Finally, as a means of fusing modalities within a LSTM, \cite{ren2016look, ma2019emotion} have separate LSTMs for each modality but they share common weights. They argue this allows the model to jointly learn correlations across modalities. 

More recently, however, transformers have been the answer to multimodal sequence-based tasks \cite{xu2022multimodal}. \cite{yu2019multimodal} captions images with a transformer composed of an image encoder to self-attend visual features and a caption decoder to generate the captions from those visual features. \cite{tsai2019multimodal} introduces cross-modal transformers, which learn the attention between each pair of modalities. The downside being that as the number of modalities increases, the number of cross-modal transformers and model size necessarily increases too. OmniNet circumvents this issue by arranging elements of a multimodal sequence into a temporal and spatial cache and then feeds both into a spatial-temporal transformer. This mechanism allows temporal features to attend over the spatial features and implicitly learn a shared representation across multiple modalities. We present the full details of this spatial-temporal transformer in \S \ref{sec:omninet}.

\subsection{Early Classification}
Initial works in early classification formulate the problem in terms of standard reinforcement learning. Specifically, \cite{liu2020finding, hartvigsen2019adaptive} use REINFORCE \cite{williams1992simple}, a standard exploration-exploitation policy gradient method to learn their early classifiers. \cite{cao2023policy} uses PPO \cite{schulman2017proximal}, another policy gradient method, to the same effect. These methods rely on trial and error and lack the ability to `look forward' to see that waiting longer for more elements would have been beneficial. In fact, \cite{cao2023policy} experimentally demonstrates that PPO performs considerably worse than such forward-looking methods like Length Adaptive Recurrent Model (LARM) \cite{huang2017length} and CIS. Accordingly, we only benchmark CIS against LARM. Again, all previous work mentioned here is with unimodal sequences. Before summarizing LARM, in the next subsection we mathematically formulate early classification and establish notation.

\subsection{Problem Setup Notation}
The set of training data $\mathcal{X}$ comprises samples $x^{(i)}$ and one-hot encoded labels $y^{(i)} \in \left\{ 0, 1 \right\}^{C}$, where $C>1$ is the number of classes and
\begin{displaymath}
x^{(i)}= \left( \left(x^{(i)}_1, m^{(i)}_1 \right), \left( x^{(i)}_2, m^{(i)}_2 \right), ..., \left( x^{(i)}_{T_\text{end}} , m^{(i)}_{T_\text{end}} \right)\right)
\end{displaymath}
are sequences of elements $x^{(i)}_t$ of modality $m^{(i)}_t$. For a sample $i$ at time $t$, its state is given by
\begin{displaymath}
s^{(i)}_t =  \left( \left(x^{(i)}_1, m^{(i)}_1 \right), \left( x^{(i)}_2, m^{(i)}_2 \right), ..., \left( x^{(i)}_{t} , m^{(i)}_{t} \right)\right).
\end{displaymath}
\emph{Classifier} neural network $f$ parameterized by $\alpha$ takes $s_t$ as input\footnote{We only explicitly write superscript samples $x^{(i)}, s^{(i)}$ when needed to distinguish.} and outputs predicted class distribution vector $\widehat{y}_\alpha \left(\cdot  | s_t \right)$. \emph{Policy} neural network $g$ parameterized by $\beta$ takes $s_t$ as input and outputs policy distribution vector $\pi_\beta \left( \cdot | s_t \right)$ over two actions (`wait' and `stop and classify now'). 
\begin{displaymath}
  \widehat{y}_\alpha \left(\cdot  | s_t \right)  = f_\alpha \left( s_t \right) 
\end{displaymath}
\begin{displaymath}
\pi_\beta \left( \cdot | s_t \right)  = g_\beta \left( s_t \right) 
\end{displaymath}
At each time step $t$, we take an action $a_t$ according to policy $\pi_\beta \left( \cdot | s_t \right)$. This action can be selected stochastically via sampling or deterministically by choosing the argmax action. We keep waiting another time step and receiving new elements $\left( x_{t+1}, m_{t+1} \right)$ until we decide to stop. Once we decide to stop and classify, we make a classification according to $ \widehat{y}_\alpha\left( \cdot  | s_t \right)$.  

To learn classifying as accurately as possible, as quickly as possible, we implement the following reward function
\begin{displaymath}
R_t^\alpha (s_t, a_t) = \begin{cases}
-\mu \quad \text{if $a_t = \text{`wait'}$} \\
-\mu - \text{CE} \left(y,   \widehat{y}_\alpha \left(\cdot  | s_t \right)  \right) \quad \text{if $a_t = \text{`stop'}$ or $t=T_\text{end}$} 
\end{cases}
\end{displaymath}
where $\mu$ is a time penalty parameter and CE is cross-entropy. Each time step incurs a constant time penalty of $-\mu$. We denote the time the model stops and classifies as time $T\leq T_\text{end}$. Early classifying can then be formulated in terms of the following optimization problem
\begin{equation}
\max_{\alpha, \beta} \E_\mathcal{X} \sum_{t} R_t^\alpha \left( s_t, a_t \left( \beta \right) \right). \label{eq:cumulativeReward}
\end{equation}
Maximizing this cumulative reward means classifying as accurately as possible (so that cross entropy is low), as quickly as possible (so that the sum of time penalties is low). The time penalty parameter $\mu$ captures how much waiting another time step is penalized. As $\mu$ grows, we may sacrifice more accuracy for earlier classifications, and vice-versa.  

\subsection{LARM}
LARM \cite{huang2017length} learns to early classify in a probabilistic manner. If $A_T$ is the decision sequence where the policy decided to stop and classify at time $T$, then this decision sequence is uniquely defined by the sequence of actions 
\begin{displaymath}
A_T = \left(a_1 = \text{`wait'}, ..., a_{T-1} =\text{`wait'}, a_T=\text{`stop and classify'} \right).
\end{displaymath}
We can also explicitly calculate the probability of decision sequence $A_T$ from policies $\pi_{\beta} \left( \cdot | s_t \right)$ as
\begin{displaymath}
\pr \left( A_T | s_T \right) = \prod_{t=1}^{T} \pi_{\beta} \left( a_t | s_t \right).
\end{displaymath}
With these stopping time probabilities, LARM seeks to maximize an expected cumulative reward based on \eqref{eq:cumulativeReward} to learn its early classifier.
\begin{displaymath}
\min_{\alpha, \beta} \E_\mathcal{X} \left[ \text{CE} \left(y,  \sum_{T=1}^{T_\text{end}}  \widehat{y}_\alpha \left( \cdot | s_T \right) \pr \left( A_T | s_T \right)  \right)  + \mu  \sum_{T=1}^{T_\text{end}} T \cdot \pr \left( A_T | s_T \right) \right] 
\end{displaymath}
The first term in this loss is a micro-averaged cross-entropy and the second term is the expected stopping time penalty. Again, for both terms the expectation is taken with respect to the stopping time $T$ probability.

We see that if $\pi_{\beta} \left( a_t  = \text{`wait'} | s_t \right)$ are small then $\pr \left( A_T | s_T \right)$ may decrease to 0 rapidly. This is tantamount to not `waiting' far enough into the sequence to gain valuable information. To prevent this, LARM sets the factors $\pi_{\beta} \left( a_t  = \text{`wait'} | s_t \right)$  to 1 with probability $\rho$ during training. This ensures the model will wait for more elements in the sequence initially. Again, we emphasize LARM is a capable of looking forward and learning when waiting will be beneficial. During inference, LARM follows a stochastic policy rollout (samples action $a_t \sim \pi_{\beta} \left( \cdot | s_t \right)$) but deterministically classifies.

\section{Spatial-temporal transformer} \label{sec:omninet}
We follow OmniNet's \cite{pramanik2019omninet} structure of first funneling elements of the multimodal sequence through their respective peripherals and then inserting those outputs in the temporal and spatial caches of the transformer segment. This section outlines this process, terminating with the policy and classifier decisions. 

\subsection{Peripherals} \label{sec:peripherals}
Before the state $s_t$ reaches the transformer block of the neural network model, modality-specific peripheral functions are applied to each element in the sequence. For instance, an image peripheral is applied to image elements and a text peripheral is applied to text elements. If there are multiple sources of text elements, we can have a separate peripheral for each. The purpose of each modality (or source's) peripheral is two-fold: First, peripherals extract relevant features. Second, peripherals project each element to a common dimension size $d_\text{model}$, a necessity for a unified transformer. Consider an image of shape $(hw,3)$ where $h,w$ are the height and width of the image in pixels and 3 refers to the RGB channels. The image peripheral will project this image to dimension $(h'w', d_\text{model})$ where $h',w' > 1$ are reduced, downsampled height and width subpixels. For text, there is no spatial dimension so we write their shape as $(1, n)$ where $n$ is the number of words or tokens. Similarly, the text peripheral will project this text to dimension $(1, d_\text{model})$.

\begin{figure}[h!]
\centerline{
\includegraphics[width=0.95\linewidth]{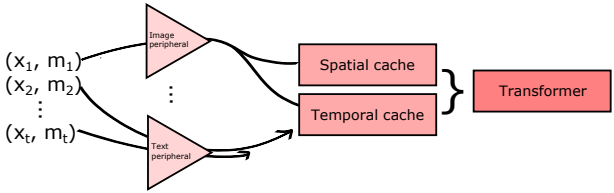} 
}
\caption{Diagram of our neural network architecture showing multimodal sequence elements passing through their respective peripherals, then being sequentially stored in the the temporal and spatial caches of the transformer.}
\label{fig:peripherals}
\Description{Flow diagram depicting connections from data to peripherals to spatial and temporal caches of transformer.}
\end{figure}

Figure \ref{fig:peripherals} depicts the overall structure of our neural network architecture with a didactic peripheral flow example. Say the first element $x_1$ of state $s_t$ has modality $m_1= \text{image}$. Note, images are the only spatial modality in our case. The image peripheral is applied and the resulting output is appended to the spatial cache. In addition, the spatial average is appended to the temporal cache. The second element $x_2$ has modality $m_2 = \text{text}$, which has no spatial dimension. We apply the text peripheral to $x_2$ and that output is sequentially appended to the temporal cache only. This goes on for all of the element in state $s_t$ with only spatial modalities (images) being appended to the spatial cache. Spatially averaged peripheral outputs of all modalities are stored in the temporal cache. Algorithm \ref{alg:peripherals} rigorously enumerates each step of this procedure.

\begin{algorithm}[h!] 
$\text{Temporal cache} = [\left. \right. ]$, $\text{Spatial cache} = [\left. \right.  ]$ \\
$d_s = \text{spatial dimensions} $ \\
\textbf{for} $t' = 1,2,...,t$ \textbf{do} \\
\qquad $\widetilde{x}_{t'} \leftarrow m_{t'} \text{ peripheral} \left( x_{t'} \right) \in \R^{d_s\times d_\text{model}}$ \\
\qquad \textbf{if} $d_s > 1 $ \textbf{do}  \\
\qquad\qquad $\text{Spatial cache} \leftarrow \left[ \text{Spatial cache}, \widetilde{x}_{t'}  \right] $ \\
\qquad \textbf{end if} \\
\qquad $\text{Temporal cache} \leftarrow \left[ \text{Temporal cache}, \frac{1}{d_s} \sum_{i=1}^{d_s} \widetilde{x}_{t'}[i, :]   \right] $ \\
\textbf{end for} 
\caption{Storing peripheral outputs in spatial and temporal caches}
\label{alg:peripherals}
\end{algorithm} 

\subsection{Transformer}
\begin{figure}[h!]
\centerline{
\includegraphics[width=0.95\linewidth]{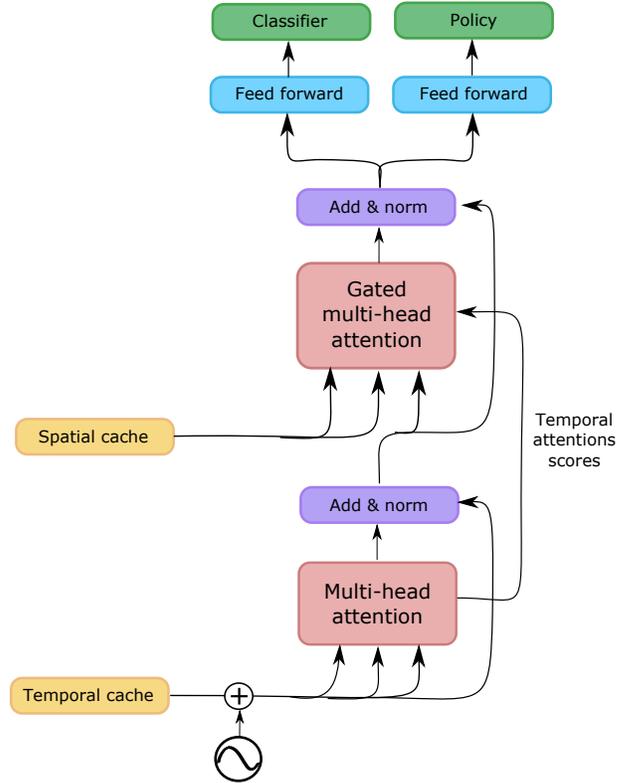} 
}
\caption{Diagram of spatial-temporal transformer body with classifier and policy heads.}
\label{fig:transformer}
\Description{Flow diagram depicting temporal and spatial caches self-attending via multi-head attention before entering as input to policy and classifier heads.}
\end{figure}

The temporal and spatial caches form the inputs into the transformer portion of the neural network. Figure \ref{fig:transformer} shows a schematic of OmniNet-based spatial-temporal transformer body with classifier and policy heads. First, the standard positional encoding \cite{vaswani2017attention} is added to the temporal cache before it enters the first multi-head attention block (with residual addition and layer normalization \cite{ba2016layer}). This first attention block's output form the `queries', and along with the `keys' and `values' from the spatial cache, make up the inputs to the \emph{gated} multi-head attention block \cite{pramanik2019omninet}. In this way, temporal features can attend to spatial features and learn complementary cross-modality information. Furthermore, with gated multi-head attention, temporal attention scores respectively scale the corresponding attention scores of spatial cache elements. For instance, if an image receives high attention in the first, temporal attention block, its corresponding subpixels will receive higher attention in this gated, spatial attention block. This is a mechanism to ensure high temporal attention translates to high spatial attention. We refer the reader to \cite{pramanik2019omninet} for further details and discussion of gated multi-head attention blocks. Finally, the output of this gated, multi-head attention block (after another residual addition and layer normalization) form the inputs for two separate, feed forward heads: one for the policy and one for the classifier. 

\section{CIS} \label{sec:cis}
Learning early classification is equivalent to maximizing the cumulative reward in \eqref{eq:cumulativeReward}. This cumulative reward can be reformulated as a function $r$ depending on label $y$, classification prediction $\widehat{y}_\alpha   \left(\cdot | s_T \right) $, and classification time $T$ given by  $r\left( y, \widehat{y}, T \right) = -\text{CE}\left( y, \widehat{y} \right) -\mu T$. An important observation is that for a fixed $\widehat{y}$ and $y$ this becomes a simple univariate function of time $T$. Utilizing this, CIS is able to learn (i) when to stop and classify and (ii) what classification to make in a more direct, supervised manner. First, CIS seeks to make the most accurate classification prediction at every time step. Second and concurrently, CIS learns the corresponding policy which yields the resulting optimized classification time. From this duality, CIS learns the ideal policy based off of its own classifications. 

The loss function is given by $\mathcal{L}_\text{CIS}  = \E_\mathcal{X} \left[  \mathcal{L}_{\widehat{y}} +  \lambda  \cdot  \mathcal{L}_{\pi}  \right]$ where
\begin{displaymath}
\mathcal{L}_{\widehat{y}}  = \frac{1}{T_\text{end}} \sum_{t=1}^{T_\text{end}} \text{CE}\left( y,   \widehat{y}_\alpha \left( \cdot | s_{t} \right) \right) 
\end{displaymath}
\begin{displaymath}
\mathcal{L}_{\pi} = \frac{1}{T_\text{end}} \sum_{t=1}^{T_\text{end}} \text{CE}\left( \widetilde{\pi}_\alpha \left( \cdot | x, t  \right) , \pi_\beta \left( \cdot | s_t \right) \right)  
\end{displaymath}
\begin{displaymath}
\widetilde{\pi}_\alpha \left( \cdot |x,t \right)  = \begin{cases}
(1, 0) \quad \text{if $t<\widetilde{T}_\alpha \left(x,y\right)$}\\
(0,1) \quad \text{if $t \geq \widetilde{T}_\alpha \left(x,y \right)$}
\end{cases} 
\end{displaymath}
\begin{displaymath}
\widetilde{T}_\alpha \left(x,y \right) = \argmax_t r\left( y, \widehat{y}_\alpha  \left(\cdot | s_{t} \right), t  \right) .
\end{displaymath}
Vector $(1,0)$ means `wait' with probability 1 and $(0,1)$ is `stop and classify' with probability 1. 
Scaling constant $\lambda$ is a hyperparameter. 

Unlike LARM, CIS does not rely on any help waiting for enough information; it is able to directly learn the optimal classification time in a supervised manner. During training $ \widetilde{\pi}_\alpha \left( \cdot | x,t \right)$ and $\widetilde{T}_\alpha$ are calculated and treated as fixed labels per minibatch update. In inference, CIS simply takes the argmax action.

\section{Experimental results} \label{sec:exp}

\subsection{Datasets and Pareto Metric}
Our first experiment is with the N24News Multimodal News Classification dataset \cite{wang2022n24news}. It consists of New York Times news articles from different categories. Each article comprises five elements. In order of appearance, they are (i) headline, (ii) abstract, (iii) image, (iv) image caption, and (v) article body. We do not need to ingest all of the elements in an article to classify its category (economy, technology, etc.).  Instead, we ingest element by element and classify the article after ingesting a minimal number of elements. We ingest the elements in the order they naturally appear in articles. To make the article body consistent in size with the other elements, however, we pad up or truncate down to 2,000 BERT \cite{devlin2018bert} tokens and further divide it into 40 elements of 50 tokens. So each article's sequence follows (headline, abstract, image, image caption, body 1, ..., body 40). The dataset contains 61,218 news articles in 24 well-balanced categories. We reserve a random 10\% of articles to be the hold-out validation set, separate from the training set. 

The second experiment is derived from the ESP Game dataset \cite{von2004labeling}. This dataset contains annotated, everyday images; each image is paired with a list of unique words describing that image. We can swap some pairings, so those images are no longer paired with their original list of words, and create the following task: Suppose you saw an image and then read the paired words one at a time. How quickly could you determine if the image and words were correctly or incorrectly paired (binary classification)? With many of the words being generic adjectives and nouns, like `blue' or `person,' the task is not trivial. We pad each word list up to 42 words, which is the largest such list, and order the words within each list by increasing uniqueness. This is done to match the design of the original ESP Game and trend of waiting longer for increasing information. The dataset contains 100,000 samples with correct and incorrect pairings evenly split. A random 10\% of samples are reserved for the hold-out validation set. Here, cross-modality learning is explicit and necessary.

Our third and final experiment makes use of industry data and application. In this real use case, we have multimodal sequences composed of four elements: (i) structured categorical data, (ii) text, (iii) a bag of images, and (iv) another bag of images. For a given sample, each element arrives sequentially but in variable order. Associated with each sample is a binary label which we attempt to predict as accurately as possible, with as few elements received. 

In reality, the features of the structured categorical data also arrive sequentially and with variable order. While we do not have access to these finer grained arrival time stamps, in consultation with subject matter experts, we mimic this process with the following procedure: We first train an XGBoost model \cite{chen2016xgboost} to classify samples' structured data only. We can then identify no-importance features (feature importance scores of 0), low importance features (scores between 0 and 0.01), and high importance features (scores greater than 0.01). The structured data is artificially made to arrive three times. For the first arrival, we set the value of each feature to `missing' with a probability according its importance. No-importance features are made `missing' with 90\% probability, low importance features with 95\% probability, and high importance features at 99\% probability. We encode `missing' by adding a dimension to the one-hot encoding of categorical features. For the second arrival, we take the first arrival and replace `missing' values with the true value with 20\% probability. Similarly for the third arrival, we do the same on the second arrival. The first arrival replaces the original structured element's sequence position. The second arrival is inserted two positions afterwards if possible, otherwise it immediately follows the first arrival. We do the same for inserting the third arrival after the second. For example, the most common sequence is (structured 1, text, structured 2, bag of images 1, structured 3, bag of images 2). In total, this dataset contains 63,030 samples with evenly split positive and negative labels. Again, we reserve a random 10\% of samples to be the validation set.

To holistically compare LARM and CIS early classifiers, we construct their Pareto frontiers. In this way we can study the complete spectrum of each method's accuracy-timeliness tradeoffs. For a specific $\mu$,  we evaluate the early classifier over the validation set and compute the mean classification time and accuracy after each training epoch. Sweeping over a range of $\mu$ values yields a set of accuracy-timeliness tradeoff points. Finally, the Pareto frontier emerges after removing all dominated points. Furthermore, we run three independent trials of this procedure to create three Pareto frontiers per method. The mean AUC of the Pareto frontiers is a holistic measure of the early classifier's accuracy-timeliness tradeoff capacity. 

\subsection{Implementation}
For all three experiments, we sweep $\mu \in \left\{ 10^{-5}, 10^{-4}, 10^{-3}, 10^{-2}, \right.$ $\left. 10^{-1} \right\}$. For CIS, we set the scaling constant $\lambda=1$. The training set is optimized by using Adam with batch size 128 until validation accuracies and mean classification times plateau. All images are resized to $224\times224$. If we pretrain a peripheral on the sequence labels, we write the accuracy in parentheses. For all peripherals, we explain the feature extractor and implement a single feed-forward layer of dimension $d_\text{model}$ for the projector (explained in \S \ref{sec:peripherals}). For the spatial-temporal transformer, each multi-head attention block has eight heads with queries, keys, and values of dimension 64. The classifier and policy heads' feed-forward networks have one hidden layer of dimension 100 with ReLU activation. 

For the N24News experiment, four separate BERT models are pretrained for each text source: Headline (73.6\% acc.), abstract (78.8\% acc.), caption (71.0\% acc.), and body portions (69.1\% acc.). We take the final hidden state of dimension 768 as the feature extraction. A ResNet-18 \cite{he2016deep} model is pretrained on the images (48.6\% acc.). The $7\times7\times512$ spatial layer before average spatial pooling is used as the feature extraction. Peripheral accuracies are in line with \cite{wang2022n24news}. For the transformer network, $d_\text{model} = 500$ and the learning rate is $10^{-5}$ for CIS and  $10^{-6}$ for LARM. Following \cite{huang2017length}, we keep LARM's waiting parameter $\rho=0.9$.

For the ESP Game image-words experiment, we cannot pretrain an image or words peripheral since both modalities are necessary to make an accurate classification. Accordingly, we simply utilize a ImageNet-pretrained ResNet-18 as the image peripheral where, again, the $7\times7\times512$ spatial layer before average spatial pooling is used as the feature extraction. For the words peripheral we use GloVe word embeddings \cite{pennington2014glove} of dimension 300. For the transformer, $d_\text{model} = 300$ and the learning rate is $10^{-5}$ for CIS and  $10^{-6}$ for LARM. LARM's waiting parameter $\rho=0.9$ again.

Finally, for our industry experiment, we do not apply a peripheral to the three structured data arrivals, just insert the projector introduced above. For reference though, a simple 500-dimensional single hidden-layer, feed-forward classifier yields a 64.8\% accuracy for the first arrival, 78.1\% accuracy for the second, and 80.1\% for the third. We pretrain a BERT model on the last 512 tokens of the text data (67.1\% acc.) and again take the final hidden state of dimension 768 as the feature extraction. For both bags of images, we pretrain separate ResNet-18 models where the average spatial pooling layer is also across images in a bag (53.6\% and 53.0\% accs.). Again, the $7\times7\times512$ spatial layers for each image before average spatial pooling is used as the feature extractions. For this transformer, $d_\text{model} = 500$ and the learning rate is $10^{-5}$ for both CIS and  LARM. LARM waiting parameter $\rho=0.9$ lead to poor results and lowering it to $0.5$ yields the best performance.\footnote{We pledge to publish our code and add a link here upon acceptance of this paper.} 

\subsection{N24News Experiment}
\begin{figure}[h!]
\centerline{
\includegraphics[width=0.95\linewidth, trim=0.3cm 0.6cm 2.0cm 2.2cm, clip]{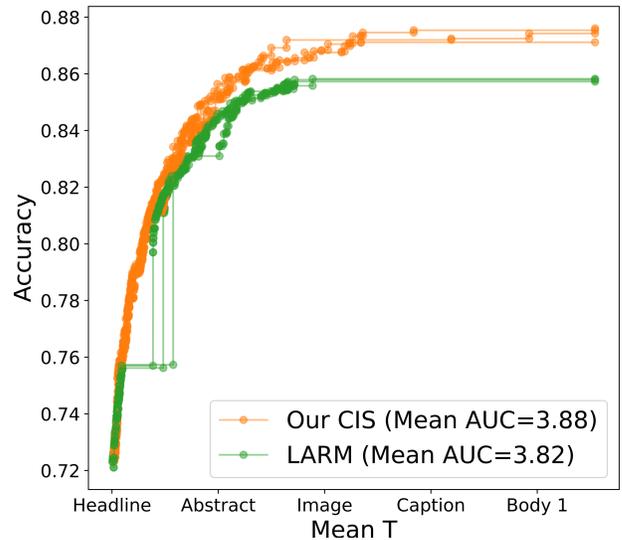} 
}
\caption{Pareto frontiers for the N24News experiment.}
\label{fig:n24NewsPareto}
\Description{Scatter plot of CIS and LARM's accuracy-mean classification time Pareto frontier points for N24News experiment.}
\end{figure}

Figure \ref{fig:n24NewsPareto} displays the Pareto frontiers for the N24News experiment. CIS's mean AUC is 1.6\% greater than LARM's mean AUC. CIS slightly outperforms LARM, and we stress this is due to the supervised nature of the algorithm. We will see this performance gap grow as the data becomes more complex and the interplay between modalities more important in the following experiments.

\subsection{ESP Game Image-Words Experiment}
\begin{figure}[h!]
\centerline{
\includegraphics[width=0.95\linewidth, trim=0.3cm 0.5cm 2cm 2.2cm, clip]{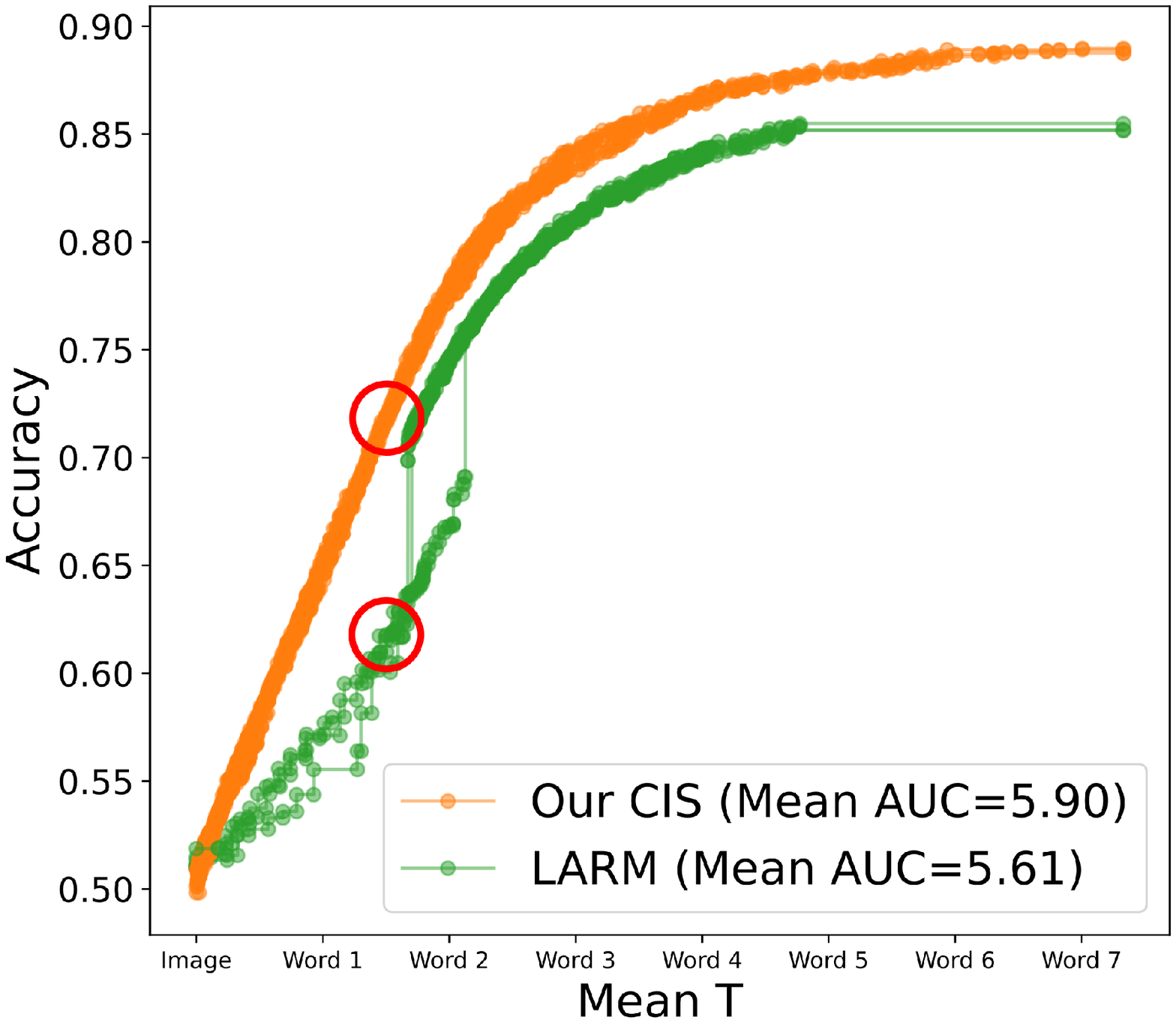} 
}
\centerline{
\includegraphics[width=0.95\linewidth, trim=0.3cm 0.3cm 1.7cm 2.2cm, clip]{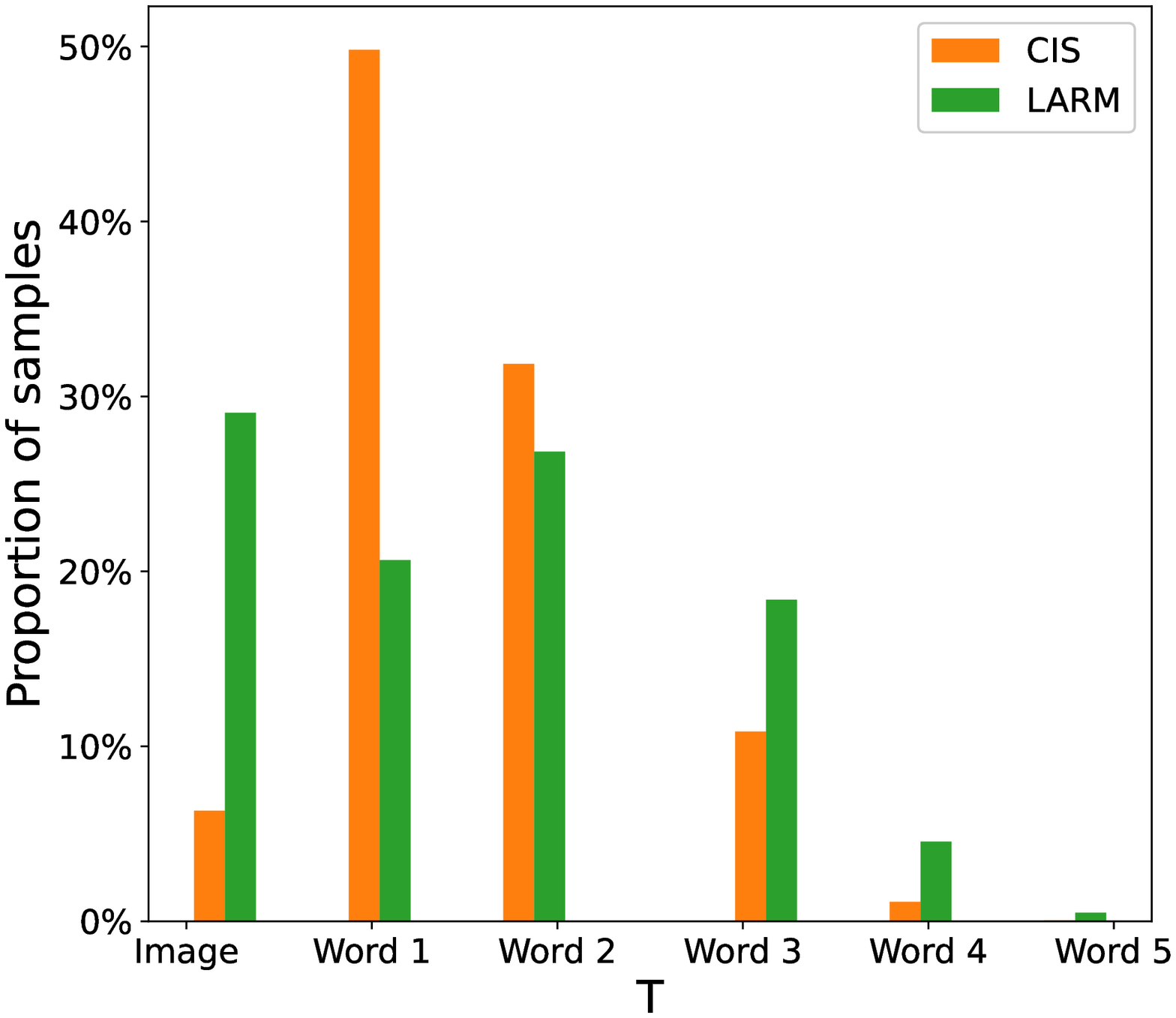} 
}
\caption{(Top) Pareto frontiers for the ESP Game image-words experiment. (Bottom) Histograms showing distribution of CIS and LARM classification times $T$.}
\label{fig:espPareto}
\Description{(Top) Scatter plot of CIS and LARM's accuracy-mean classification time Pareto frontier points for ESP Game image-words experiment. (Bottom) Bar graph showing proportion of samples that stopped at each element number for both CIS and LARM.}
\end{figure}

Figure \ref{fig:espPareto} (top) displays the Pareto frontiers for the ESP Game image-words experiment. CIS holistically outperforms LARM. CIS's mean AUC is 5.2\% greater than LARM's mean AUC. Reflected in this larger AUC margin, CIS is able to better capture the multimodal dependency.

To showcase CIS's discerning patience, we investigate the distribution of stopping times compared to LARM. Figure \ref{fig:espPareto} (bottom) shows just this using CIS and LARM with mean classification time of 1.5 words (red circles in Figure \ref{fig:espPareto} (top)). We can see that CIS (i) waits until at least the first word as that's the minimum information needed to predict accurately and (ii) does not wait for as many words as LARM on the tail end. In other words, the distribution has lower spread.

\subsection{Industry Experiment}
\begin{figure}[h!]
\centerline{
\includegraphics[width=0.95\linewidth, trim=0.3cm 0.6cm 2cm 2.2cm, clip]{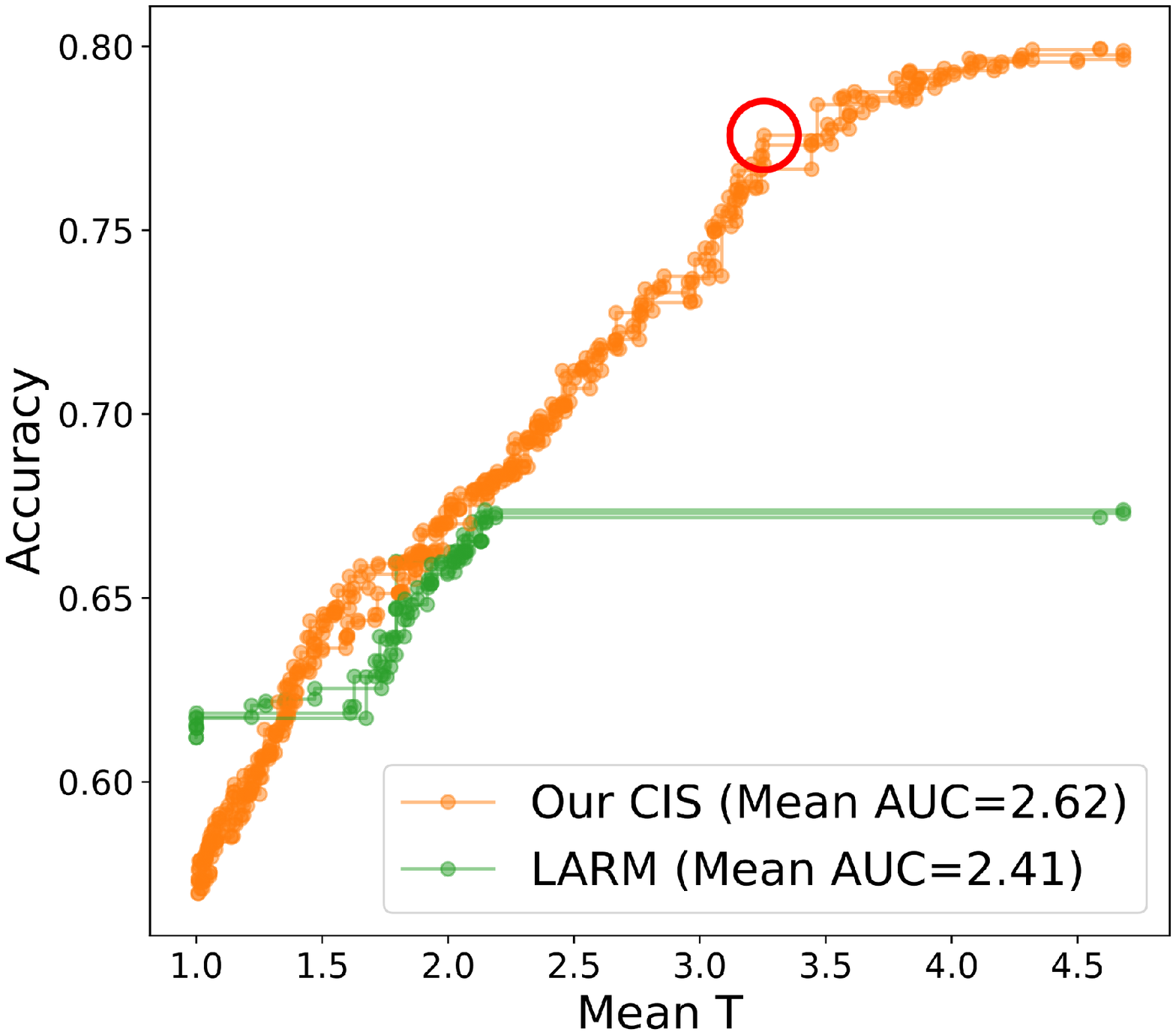} 
}
\centerline{
\includegraphics[width=0.95\linewidth, trim=1.5cm 0cm 1.9cm 1.6cm, clip]{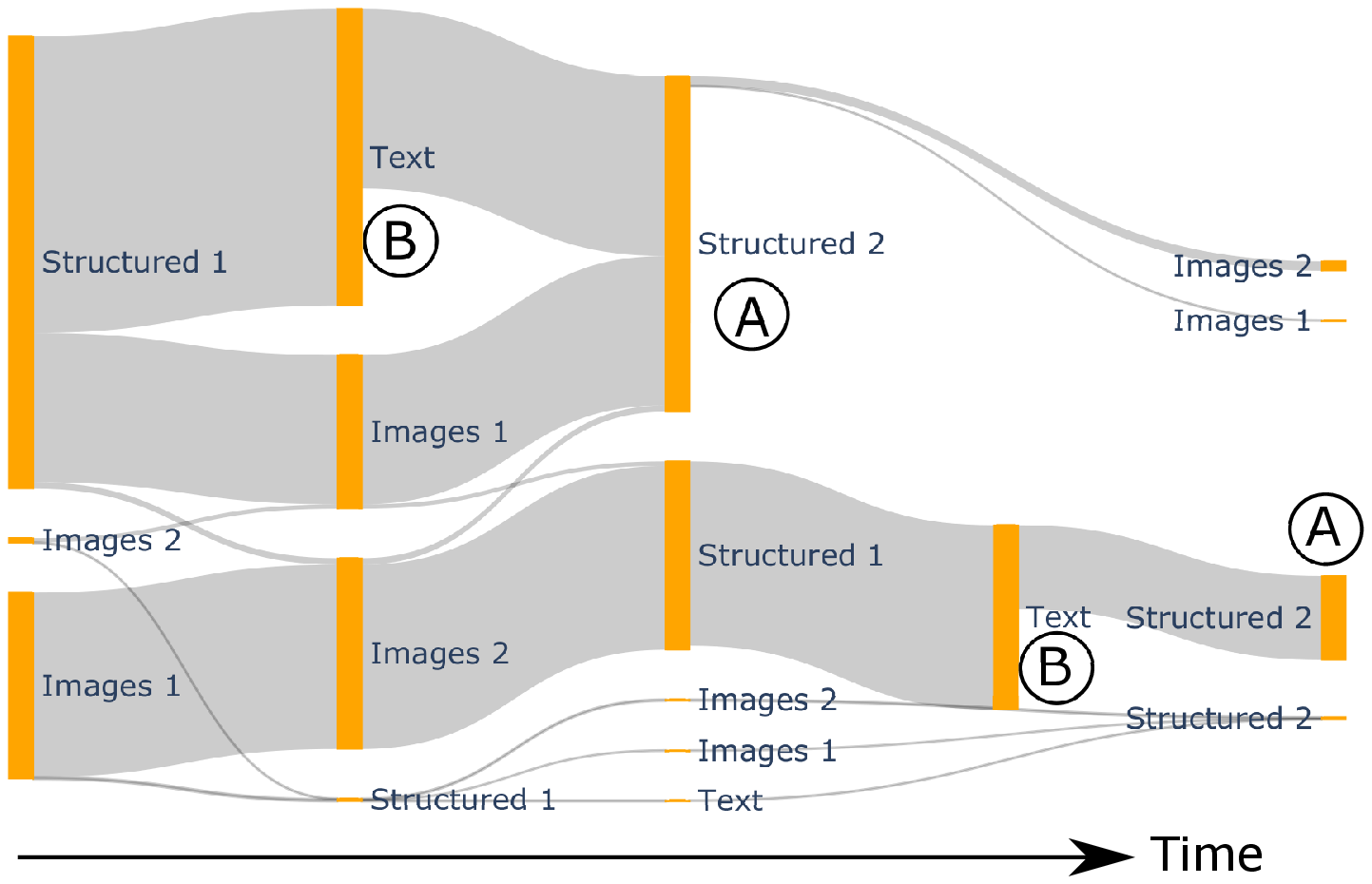} 
}
\caption{(Top) Pareto frontiers for the industry experiment. (Bottom) Sankey plot showing stopping times of samples with different modality arrivals.}
\label{fig:industryPareto}
\Description{(Top) Scatter plot of CIS and LARM's accuracy-mean classification time Pareto frontier points for industry experiment. (Bottom) Flow diagram showing when the CIS stops for samples of different modality orders.}
\end{figure}

Figure \ref{fig:industryPareto} (top) displays the Pareto frontiers for the industry experiment. Again, CIS holistically outperforms LARM. CIS's mean AUC is 8.7\% greater than LARM's mean AUC. 

We wish to study the stopping times for the variable modality arrivals. In Figure \ref{fig:industryPareto} (bottom), we show a Sankey plot of sequences and their respective stopping times for a specific CIS Pareto point (circled in red in Figure \ref{fig:industryPareto} (top)). The first observation is that CIS most often stops after receiving the second structured data (seen from `A' markers). This of course makes sense since we are in the lower time penalty region and structured data is the most informative (highest accuracy peripheral). A second, more interesting observation is that CIS also frequently stops and classifies after receiving both the first structured data and text modality in that order (seen from `B' markers). This suggests these two modalities have complementary information. As we can see, there is rich opportunity for studying early classification models. 

\section{Conclusion} \label{sec:conclude}
Early classification has recently gained attention as an important adaptation of classification to dynamic environments. Methods like LARM and CIS represent the state-of-the-art. However, these methods have solely focused on unimodal sequences. To our knowledge, this paper is the first study of early classification of multimodal sequences. Not only do we stress the ubiquity of such problems in the real world but also demonstrate that an OmniNet-like spatial-temporal transformer combined with CIS is an effective approach. For sure, multimodal sequences are an important extension of unimodal early classification.

\bibliographystyle{ACM-Reference-Format}
\bibliography{multimodalEarlyClassify}

\end{document}